\documentclass{article}

\PassOptionsToPackage{numbers}{natbib}
\usepackage{preprint}

\usepackage[utf8]{inputenc}
\usepackage[T1]{fontenc}
\usepackage{hyperref}
\usepackage{url}
\usepackage{booktabs}
\usepackage{amsfonts}
\usepackage{amsmath}
\usepackage{nicefrac}
\usepackage{microtype}
\usepackage{xcolor}
\usepackage{pifont}
\usepackage{graphicx}
\usepackage{multirow}
\usepackage{array}
\usepackage{tabularx}
\usepackage{enumitem}
\usepackage{subcaption}
\usepackage[capitalise,noabbrev]{cleveref}

\newcommand{\cmark}{\ding{51}}
\newcommand{\xmark}{\ding{55}}
\newcommand{\modelname}{Qwen3-VL-8B-Thinking}
\newcommand{\region}[1]{\mathcal{R}_{\mathrm{#1}}}
\newcommand{\attn}[1]{A_{\mathrm{#1}}}
\newcommand{\dgar}{\mathrm{DGAR}}
\newcommand{\probd}{\Delta p}
\newcommand{\tstar}{t^{*}}
\newcommand{\recovery}{\mathcal{S}}
\newcommand{\layervar}{\ell}
\newcommand{\headvar}{h}
\newcommand{\prob}{p}

\title{Do Vision–Language Models Understand 3D Scenes or Just Catalogue Objects?}

\author{%
  Animesh Maheshwari\thanks{Research lead} \\
  Deccan AI\\
  \texttt{animeshmaheshwari@deccan.ai} \\
  \And
  Divyansh Sahu \\
  Deccan AI \\
  \texttt{divyansh@deccan.ai} \\
  \AND
  Nishit Verma \\
  Deccan AI \\
  \texttt{nishit@deccan.ai} \\
}

\begin{document}

\maketitle


\begin{abstract}
Vision–language models reliably name objects in a scene, but do they represent the 3D layout those objects inhabit? We introduce a 3,034-sample human-curated benchmark targeting three components of spatial understanding: depth-ordered occlusion (probed via three independent counterfactual operationalisations), optical-geometry inference over visible reflections, and volumetric rearrangement planning. Six frontier and open-weight VLMs, scored by trained annotators on 18,204 responses with no LLM-as-judge, reveal a sharp dissociation: models that plan rearrangements over visible layouts at 53–97\% accuracy and rarely violate collision constraints fall to 6–45\% on occlusion and below 7\% on reflections. An embodied-reasoning model reproduces the same profile. White-box analysis on Qwen3-VL-8B-Thinking localises the failure to the visual-token merger: spatial information recoverable throughout the vision encoder becomes inaccessible after token compression and only stabilises again when clean post-merger activations are patched into the language decoder. 
\end{abstract}


\section{Introduction}
\label{sec:intro}

A system that registers only what is directly visible sees a stack of plates; a system that understands the scene infers that a mug occluded behind them persists, occupies a particular volume, and lies at a reachable depth. The second capability, reasoning about what the image \emph{does not show} but the scene nonetheless contains, is what allows perception to extend beyond the pixel array. It requires an internal \emph{3D structured scene model}: a representation in which objects have depth and extent, occlusion is a geometric relation rather than an absence, and the 2D image is understood as one projection of a richer underlying scene.

Vision--language models are increasingly positioned as the perceptual layer for systems where this distinction is consequential, and downstream components inherit whatever structure the VLM provides~\cite{brohan2023rt2,driess2023palme,ha2018worldmodels,bardes2024vjepa}. A growing body of work has begun probing VLMs beyond static recognition, including counterfactual and ``what-if'' style evaluations targeting causal reasoning, physical plausibility, and commonsense inference (\Cref{tab:related_comparison}). Our focus is narrower and complementary: we ask whether VLMs maintain the \emph{geometric} substrate that such reasoning, in the spatial domain, would seem to require. Concretely, we study two competencies that we argue are diagnostic of an internal 3D scene model rather than a 2D catalogue of detections: \emph{spatial counterfactual reasoning}, the ability to predict what an intervention on the scene would expose or alter at specific 3D locations, and \emph{optical-geometry inference}, the ability to trace projected image regions (shadows, reflections, occlusion boundaries) back to the 3D objects that produce them. The first overlaps in surface form with prior counterfactual benchmarks but is grounded in spatial intervention rather than event-level outcomes; the second has, to our knowledge, received less direct attention.

This paper asks whether modern VLMs maintain such a representation, and answers along three converging axes: a behavioural axis, through a benchmark probing spatial counterfactual reasoning in three independent ways; an attentional axis, through a depth-grounded attention metric; and a causal axis, through activation patching.

\paragraph{What the data shows.} A clean dissociation emerges, but it is not the binary of visible-vs-hidden; it is whether the question is answerable from a flat object catalogue or requires 3D scene structure. Volumetric planning, where the structure is essentially \emph{given} by the layout, is solved at 53--97\% by six VLMs spanning proprietary and open-weight systems, dense and MoE architectures with CoT reasoning, and three orders of magnitude in active parameters; collision-feasibility errors stay under 7\%. Occlusion reasoning, where the structure must be \emph{inferred} from depth ordering, collapses to 6--45\%, and optical geometry, where it must be \emph{traced} across cross-region projection, collapses to 1--7\%. Holding scenes, prompting, and scoring constant, the same six models exhibit a 50+ point gap between competencies that test the same physical scene through different demands on internal 3D structure. We probe occlusion three independent ways (deletion, minimum-set deletion, volumetric transparency); their convergence within a 12-point band for five of six models indicates the missing piece is the underlying 3D representation, not any task-specific surface feature.

\paragraph{From behaviour to mechanism.} On Qwen3-VL-8B-Thinking ~\cite{qwen3vl8bthinking}, the only open-weight VLM in our sweep small enough to instrument with full activation access on a single GPU, our \emph{Depth-Guided Attention Relevance} (DGAR) metric shows visual attention to the geometrically correct image region is dispersed and at-or-below uniform chance during failures. Causal tracing via activation patching localises the failure to the visual-token merger: spatial information that is recoverable inside the ViT becomes inaccessible to downstream computation after compression, and only re-stabilises at the first language-model layer when post-merger activations are restored. We frame this as a single-model finding about an architectural pattern shared by current ViT-plus-LM VLMs rather than a universal claim, and discuss what would falsify it in larger or differently architected models in \Cref{sec:discussion}.

\paragraph{Contributions.}
(i)~We introduce \emph{spatial counterfactual reasoning} as an evaluation axis for VLMs, complementing visible-state spatial reasoning, and show that three independent operationalisations of occlusion reasoning converge on a consistent failure pattern across six models, closing off task-specific or prompting-specific explanations.
(ii)~The 3{,}034-sample benchmark is scored end-to-end by trained annotators on 18{,}204 responses with full QC review; no LLM-as-judge.
(iii)~DGAR + activation patching identify the visual-token merger as a candidate architectural site where spatial structure becomes inaccessible to the language decoder, in one open-weight VLM.
(iv)~The benchmark, scoring script, and mechanistic-analysis code (DGAR computation and activation-patching pipeline) are publicly released. Per-image segmentation masks are not shipped; users generate masks for the target object in each image using SAM~3 \cite{carion2025sam3segmentconcepts}.

\paragraph{Scope.} We evaluate VLMs (image+text $\to$ text) rather than vision--language--action models (VLAs); isolating the VLM layer produces a cleaner characterisation of the perception--reasoning bottleneck VLAs inherit from their backbones. The behavioural deficit appears in all six models we test, and a targeted evaluation of \textbf{Gemini Robotics-ER 1.6}~\cite{gemini_robotics_team_2025} on a 200-sample subset of Gemini~3.1's failures reproduces the same competency profile, indicating that the deficit is not closed by current embodied-reasoning post-training. Whether the mechanistic locus we identify in one model generalises to the others is an open question (\Cref{sec:discussion}).


\section{Related Work}
\label{sec:related}

Existing spatial-reasoning benchmarks~\cite{liu2023vsr,cheng2024spatialrgpt,chen2024spatialvlm,chen2025internspatial,kamath2025mindgap,ranasinghe2024mindgap,tong2024vlmsblind,yang2024vsibench,chen2024ego3d,qian2024nuscenesspatial,lin2025ivispar} evaluate competencies over the \emph{currently visible} scene state (relation classification, region grounding, distance estimation) without probing spatial counterfactuals or optical-geometry inference. MindCube~\cite{wang2025mindcube} targets viewpoint-transformation and mental-mapping from limited views; our benchmark differs by focusing on \emph{active intervention} (removing, transparentising, rearranging an object), which demands volumetric understanding that limited-view reconstruction does not. Counterfactual benchmarks have addressed \emph{temporal} counterfactuals over video~\cite{mahmood2025countervqa,wang2025cover} and \emph{causal} counterfactuals over physics simulations or structured causal models~\cite{chen2022comphy,zhang2025causalvlbench}, but not the \emph{spatial} static-scene counterfactuals we introduce. On methodology, activation patching and causal tracing have localised factual recall in language models~\cite{meng2022locating} and modality-bridging in multimodal systems; we apply the same toolkit to a behavioural deficit. We also eschew automated evaluation~\cite{zheng2024judging,li2023alpacaeval,yu2024mmvet,liu2025mmbench} in favor of end-to-end human evaluation, since 3D-structural questions require fine-grained visual inspection that cannot be reliably delegated to a model exhibiting similar deficits. A detailed comparison and per-category discussion appear in Appendix~\ref{sec:appendix_related_table}.


\section{Benchmark Design}
\label{sec:benchmark}

Our benchmark consists of 3{,}034 samples organised around a single scientific question: do VLMs maintain the 3D structured scene model that physical-world reasoning requires, or only a flat 2D catalogue of detected objects? Existing benchmarks cannot answer this because they evaluate models against the scene that is currently visible. We instead ask the model to reason about state that is \emph{not} displayed: predicting what would become visible under an intervention, or tracing a visible reflection back to the 3D object that produces it. We refer to the first class as \emph{spatial counterfactual reasoning}.

The six tasks span three competencies (\Cref{tab:benchmark_overview}). \textbf{Depth-ordered occlusion (T1, T2, T3)} probes whether the model encodes which objects sit in front of which, through three independent counterfactual operationalisations differing in surface phrasing, in the direction of the counterfactual (deletion vs.\ transparency), and in the inference required (pairwise depth ordering, combinatorial set selection, volumetric ray-casting). \textbf{Optical geometry (T4)} probes cross-region 3D-to-2D projection: identifying which objects produce reflections visible on a surface. The reflections are present in the image, so this is not a hidden-state question but a 3D-projection question. \textbf{Volumetric planning (T5, T6)} asks whether the model can use 3D structure already given by the visible layout, by computing rearrangements that respect collision constraints, and serves as a diagnostic floor: any failure here would imply a more fundamental 3D deficit than the inference failures T1--T4 test. Extended motivation for each task and the design principles underlying them appear in Appendix~\ref{sec:appendix_competencies_extended}.

All images are synthetically generated with Nano Banana 2 at 720$\times$480, tuned for photorealistic indoor composition, and human-validated for occlusion structure and depth gradation. Synthetic generation is a requirement, not a convenience: the benchmark needs scenes with controlled occluder counts, calibrated depth gradients, unique-referent labels, and per-scene counterfactual ground truth (what would be revealed if X were removed), a combination no existing real-world dataset (COCO, Visual Genome, ScanNet, Hypersim) provides. We mitigate the residual risks (rendering artefacts, drift from natural images) via per-image human validation and a planned real-photograph replication of T1 and T4 (\Cref{sec:discussion}). Ground-truth annotations and response evaluations are produced entirely by trained human annotators (15 annotators + 5 QC reviewers, every sample QC-reviewed). Each response is graded as Correct, Incorrect, or Hallucinated with no partial credit; we adopt the inclusive hallucination definition of~\cite{rohrbach2018object,wang2023amber}. All prompts use natural-language noun phrases with no bounding boxes, masks, or coordinates; supplying coordinates would short-circuit the linguistic-to-spatial grounding step we are testing. Construction details, canonical visibility/occlusion definitions, the full annotation protocol, and prompt templates appear in Appendices~\ref{sec:appendix_definitions}--\ref{sec:appendix_prompts}.

\begin{table*}[t]
\centering
\footnotesize
\setlength{\tabcolsep}{3pt}
\renewcommand{\arraystretch}{1.05}
\caption{\textbf{Task taxonomy.} T1--T3: spatial counterfactual probes of depth-ordered occlusion (three independent operationalisations of the same competency). T4: optical-geometry probe over a visible reflection. T5--T6: rearrangement probes with volumetric understanding over fully visible state, serving as a diagnostic control.}
\label{tab:benchmark_overview}
\begin{tabularx}{\textwidth}{l l c X X X X}
\toprule
\textbf{Competency} & \textbf{Task} & \textbf{$N$} & \textbf{Prompt} &
\textbf{Ground Truth} & \textbf{Correct Criterion} & \textbf{Core Demand} \\
\midrule
\multirow{4}{*}{\shortstack[l]{\textit{Depth-ordered}\\\textit{occlusion}}}
& T1: Single-Object Removal & 602 & Remove \{X\}; identify newly visible object(s)
  & Objects occluded exclusively by X & Complete recall, no fabrication
  & Pairwise depth ordering \\
\cmidrule(l){2-7}
& T2: Multi-Object Removal   & 330 & Minimum removals to reveal \{X\}
  & Smallest valid set ($\geq$2) & Exact set
  & Combinatorial occlusion-chain inference \\
\cmidrule(l){2-7}
& T3: Transparency    & 900 & Make \{X\} transparent; identify visible objects
  & Directly behind X only & Complete recall
  & Volumetric ray-casting through X's volume \\
\midrule
\textit{Optical}
& T4: Reflection      & 300 & Identify reflected objects
  & Structural reflections only & Complete recall
  & Cross-region 3D-to-2D projection tracing \\
\midrule
\multirow{2}{*}{\shortstack[l]{\textit{Volumetric}\\\textit{planning}}}
& T5: 1-Swap          & 602 & One swap to match target order
  & Unique valid swap & Correct swap
  & Single-step collision-aware planning \\
\cmidrule(l){2-7}
& T6: Multi-Swap      &  300 & Multi-step swaps to match target
  & Verified sequence / infeasible & Correct swaps
  & Multi-step collision-aware planning \\
\midrule
& \textbf{Total} & \textbf{3{,}034}  & & & & \\
\bottomrule
\end{tabularx}
\end{table*}


\section{Experimental Setup}
\label{sec:experimental_design}

We evaluate six VLMs spanning proprietary and open-weight systems and three orders of magnitude in active parameters: \textbf{GPT-5.2}~\cite{gpt52}, \textbf{Gemini~3.1}~\cite{gemini31}, \textbf{GLM-4.6V}~\cite{glm46v}, \textbf{Qwen3.5-VL-397B-A17B}~\cite{qwen35vl}, \textbf{Qwen3-VL-30B-A3B-Thinking}~\cite{qwen3vl_thinking}, and \textbf{\modelname{}}~\cite{qwen3vl8bthinking}. The sweep spans dense and MoE architectures with CoT reasoning. All models support thinking-mode reasoning, which we enable. Models are queried via OpenRouter at temperature 0, single inference pass per sample, no system prompt, no few-shot examples; prompts are fixed per task (Appendix~\ref{sec:appendix_prompts}). Mechanistic experiments (\Cref{sec:mechanistic}) run on \modelname{}, the only open-weight VLM in our sweep small enough to instrument with full activation access on a single A100, over 163 failure cases (56 each for T1 and T3, 51 for T2). T4--T6 are excluded from aggregate mechanistic analysis; \textbf{T4 appears as a visual grounding case study} in Appendix~\ref{sec:appendix_reflection}. All experiments run on a single NVIDIA A100 80\,GB GPU.


\section{Results}
\label{sec:experiments}

\Cref{tab:main_results} reports correctness rates across all six models and tasks. The results table is structured as follows: a model's three scores on T1/T2/T3 reveal whether it has depth-ordered occlusion structure; its T4 score reveals whether it has optical-geometry tracing; its T5/T6 scores reveal whether it can use 3D structure already visible.

\begin{table}[t]
\centering
\caption{\textbf{Correctness (\%) across models, organised by 3D-scene-structure competency.} The two depth-inference competencies (occlusion, optical) collapse uniformly; volumetric planning succeeds. The boundary between competencies is sharper than any difference within them. Best per task in \textbf{bold}; second-best \underline{underlined}.}
\label{tab:main_results}
\resizebox{\linewidth}{!}{%
\begin{tabular}{l|ccc|c|cc}
\toprule
& \multicolumn{3}{c|}{\textbf{Depth-ordered occlusion}} & \textbf{Optical} & \multicolumn{2}{c}{\textbf{Volumetric planning}} \\
\textbf{Model} & \textbf{T1} & \textbf{T2} & \textbf{T3} & \textbf{T4}
& \textbf{T5} & \textbf{T6} \\
& \small Single-Obj.\ Rem. & \small Multi-Obj.\ Rem. & \small Transp. & \small Refl.
& \small 1-Swap & \small Multi-Swap \\
\midrule
GPT-5.2                 & \underline{23.01 $\pm$ 3.48} & 24.24 $\pm$ 4.85 & 24.08 $\pm$ 2.89 & \underline{4.00 $\pm$ 2.33} & 92.52 $\pm$ 2.49 & 77.33 $\pm$ 5.33 \\
Gemini~3.1              & \textbf{44.37 $\pm$ 4.14} & \textbf{44.24 $\pm$ 5.45} & \textbf{43.51 $\pm$ 3.28} & \textbf{7.00 $\pm$ 3.00} & \textbf{96.51 $\pm$ 1.99} & \underline{90.00 $\pm$ 4.00} \\
Qwen3.5-VL-397B-A17B    & 22.85 $\pm$ 3.48 & \underline{35.76 $\pm$ 5.45} & \underline{24.75 $\pm$ 2.89} & 2.00 $\pm$ 1.67 & \underline{94.85 $\pm$ 2.33} & \textbf{93.33 $\pm$ 3.67} \\
Qwen3-VL-30B-A3B-Thinking            & 21.03 $\pm$ 3.31 & 13.94 $\pm$ 3.94 & 15.32 $\pm$ 2.44 & 2.33 $\pm$ 1.67 & 80.73 $\pm$ 3.65 & 67.33 $\pm$ 5.67 \\
GLM-4.6V                & 14.57 $\pm$ 2.81 &  9.09 $\pm$ 3.03 & 11.10 $\pm$ 2.11 & 1.67 $\pm$ 1.67 & 77.41 $\pm$ 3.65 & 72.33 $\pm$ 5.67 \\
\modelname{}            & 14.90 $\pm$ 2.81 &  8.79 $\pm$ 3.03 &  6.44 $\pm$ 1.66 & 2.67 $\pm$ 2.00 & 67.61 $\pm$ 3.99 & 53.33 $\pm$ 6.00 \\
\bottomrule
\end{tabular}%
}
\end{table}

\paragraph{Occlusion reasoning fails on three independent probes.}
T1, T2, and T3 each ask the model to reason about objects whose relationship to the foreground is determined by depth order, but they differ in surface phrasing, the direction of the counterfactual, and the type of inference. If the deficit lay in any of these surface attributes, the three would dissociate. They do not. For five of six models, accuracies on T1, T2, and T3 fall within a 12-point band, an extraordinarily tight clustering for tasks with such different surface forms. Gemini~3.1 scores 44.37/44.24/43.51\%; GPT-5.2 scores 23.01/24.24/24.08\%. Qwen3.5-VL-397B-A17B is the only model where the three probes diverge meaningfully (22.85/35.76/24.75\%), and even there the spread is bounded by 13 points. The relative ordering of models is also stable across all three probes. Four alternative explanations are closed off: prompt-specific failure (different templates), operationalisation-specific failure (deletion, set-inference, transparency all collapse together), difficulty-specific failure (the best model on T1 still sits 52 points below its own T5 ceiling), and scale-specific failure (moving from Qwen3-VL-8B-Thinking through Qwen3-VL-30B-A3B-Thinking to Qwen3.5-VL-397B-A17B raises T1 only from 14.90\% to 21.03\% to 22.85\% across roughly a 50$\times$ span in total parameters). What remains is a competency-level deficit: VLMs do not maintain the depth-ordered scene representation occlusion reasoning requires.

\paragraph{Reflections expose a distinct failure mode at the floor of accuracy.}
T4 sits at the floor of model performance: no model exceeds 7\%, and three are below 3\%. The reflections in T4 are \emph{not hidden}; they are fully present in the image, so failure cannot be attributed to occluded state. The qualitative case study in Appendix~\ref{sec:appendix_reflection} shows that when the model fails, it never localises reflections \emph{as such} but instead emits objects from a learned ``shiny surfaces show reflections'' prior. We treat T4 as a complementary probe whose qualitative failure mode (cross-region projection tracing) is what the data supports, rather than as a precise measurement of how often it occurs.

\paragraph{Volumetric planning works, and rules out the obvious confounds.}
Five of six models exceed 67\% on T5 and three exceed 92\%; on T6, three models exceed 70\% even with multi-step search. The strongest objection to the dissociation thesis is that planning failures might \emph{also} be 3D-structural: models fail at swaps because they cannot judge collisions. We measured this directly. Across all six models and both planning tasks, volumetric-feasibility errors account for 0.00\%--6.70\% of total samples (\Cref{tab:vol_issues}); when the 3D structure is given, models compute collision constraints correctly, and the planning failures that do occur are planning-side (search-budget exhaustion, off-by-one swap counts, mis-ordered sequences). Holding scenes, prompting, and scoring constant, the same six models score 6--45\% when 3D structure must be \emph{inferred}, 1--7\% when it must be \emph{traced} across optical projection, and 53--97\% when it is already \emph{visible}. The boundary is not visible-vs-hidden, since reflections in T4 are fully visible; it is whether the question can be answered from a flat object catalogue. The gap persists across three orders of magnitude in active parameters and with thinking-mode reasoning enabled, indicating that the bottleneck is upstream of reasoning: a representational rather than an inferential limit, which the next section investigates directly.

\begin{table}[t]
\centering
\caption{\textbf{Volumetric-feasibility error rates on the planning family.} Percentage of total samples per task in which a model proposes a swap that would cause two objects to physically overlap. Across both tasks and all six models, volumetric errors stay under 7\%: when the 3D structure is given by visible layout, models perceive object volumes correctly.}
\label{tab:vol_issues}
\small
\begin{tabular}{lcc}
\toprule
\textbf{Model} & \textbf{T5: 1-Swap (\%)} & \textbf{T6: Multi-Swap (\%)} \\
\midrule
GPT-5.2                 & 0.20 & 2.70 \\
Gemini~3.1              & 0.00 & 2.30 \\
Qwen3.5-VL-397B-A17B    & 0.00 & 2.30 \\
Qwen3-VL-30B-A3B-Thinking            & 0.70 & 4.70 \\
GLM-4.6V                & 1.30 & 3.30 \\
\modelname{}            & 2.00 & 6.70 \\
\bottomrule
\end{tabular}
\end{table}

\paragraph{Embodied-reasoning post-training does not close the gap.}
We evaluated \textbf{Gemini Robotics-ER 1.6}~\cite{gemini_robotics_team_2025}, a model from the same family as our strongest baseline that has been post-trained for embodied spatial reasoning, on a 200-sample subset drawn entirely from items Gemini~3.1 got wrong (50 per task across T1--T4). \Cref{tab:gemini_er} reports an overall 14.0\% (T1: 18.0\%, T2: 22.0\%, T3: 14.0\%, T4: 2.0\%), the same competency profile as the general-purpose VLMs in \Cref{tab:main_results}. Embodied post-training shifts absolute numbers but leaves the structural pattern intact.

\begin{table}[h]
\centering
\caption{\textbf{Gemini Robotics-ER 1.6 on a 200-sample subset of Gemini~3.1's failures} (50 per task, T1--T4). Accuracy reflects performance on items previously failed by Gemini~3.1. The ER-trained model exhibits similar competency profile as general-purpose VLMs}
\label{tab:gemini_er}
\small
\setlength{\tabcolsep}{5pt}
\begin{tabular}{lccc}
\toprule
\textbf{Task} & \textbf{Correct} & $N$ & \textbf{Accuracy} \\
\midrule
T1: Single-Object Removal       & 9  & 50  & 18.0\% \\
T2: Multi-Object Removal & 11 & 50  & 22.0\% \\
T3: Transparency  & 7  & 50  & 14.0\% \\
T4: Reflection    & 1  & 50  & \phantom{0}2.0\% \\
\midrule
\textbf{Overall}  & \textbf{28} & \textbf{200} & \textbf{14.0\%} \\
\bottomrule
\end{tabular}
\end{table}


\section{Mechanistic Analysis}
\label{sec:mechanistic}

We now ask: \emph{where} in the model's representations does spatial reasoning break down, and \emph{why}? We conduct two complementary studies on \modelname{}, focusing exclusively on failure cases across T1, T2, and T3 (163 total: 56 each for T1 and T3, 51 for T2). Both share a common spatial-grounding pipeline: per-image monocular depth via Depth-Anything-V3~\cite{depthanything3} and per-object segmentation via SAM~3~\cite{carion2025sam3segmentconcepts}. Implementation details are in Appendix~\ref{sec:appendix_impl}.

\subsection{Depth-Guided Attention Relevance (DGAR)}
\label{sec:dgar}

DGAR measures whether the model's attention, at the moment of answer generation, focuses on the \emph{depth-correct} region: the spatial neighbourhood where answer objects reside according to scene geometry. We partition each example's visual token grid into three disjoint subsets: \textbf{Target ($\region{T}$)}, tokens overlapping the query-object mask; \textbf{Depth-correct ($\region{D}$)}, tokens at the geometrically correct depth plane within an expanded bounding box of $\region{T}$ (shallower than the target for T2, deeper for T1 and T3); and \textbf{Irrelevant ($\region{I}$)}, all remaining reliable visual tokens. Aggregating last-position attention as $\attn{R} = \sum_{i \in \region{R}} \alpha_i^{(\layervar,\headvar)}$ with standard scaled-dot-product softmax, we define
\begin{equation}
  \dgar(\layervar, \headvar)
  = \frac{\attn{D}}{\attn{T} + \attn{D} + \attn{I}},
  \qquad
  \dgar_{\mathrm{chance}}
  = \frac{|\region{D}|}{|\region{T}| + |\region{D}| + |\region{I}|}.
  \label{eq:dgar}
\end{equation}
We additionally track target fixation $\mathrm{TF} = \attn{T}/(\attn{T}+\attn{D}+\attn{I})$ and irrelevant fraction $\mathrm{IRR} = \attn{I}/(\attn{T}+\attn{D}+\attn{I})$, and assign each failure to one of three modes via winner-take-all on per-example means: \emph{Target Fixation}, \emph{Depth-Aware but Wrong}, or \emph{Attention Dispersed}. Full region definitions, depth margins, and adaptive fallback rules are in Appendix~\ref{sec:appendix_impl}.

\begin{figure}[t]
  \centering
  \includegraphics[width=0.49\textwidth]{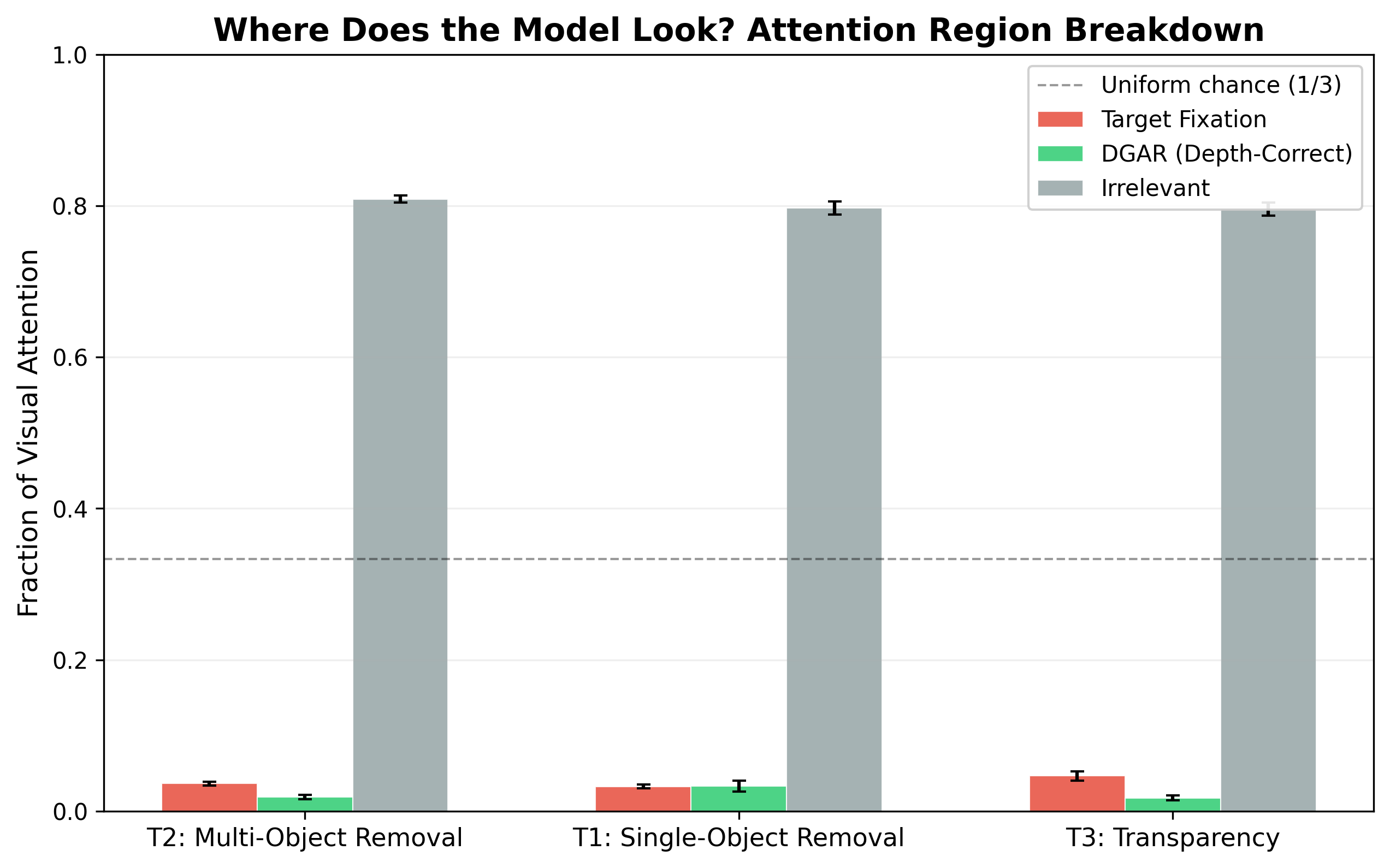}
  \hfill
  \includegraphics[width=0.49\textwidth]{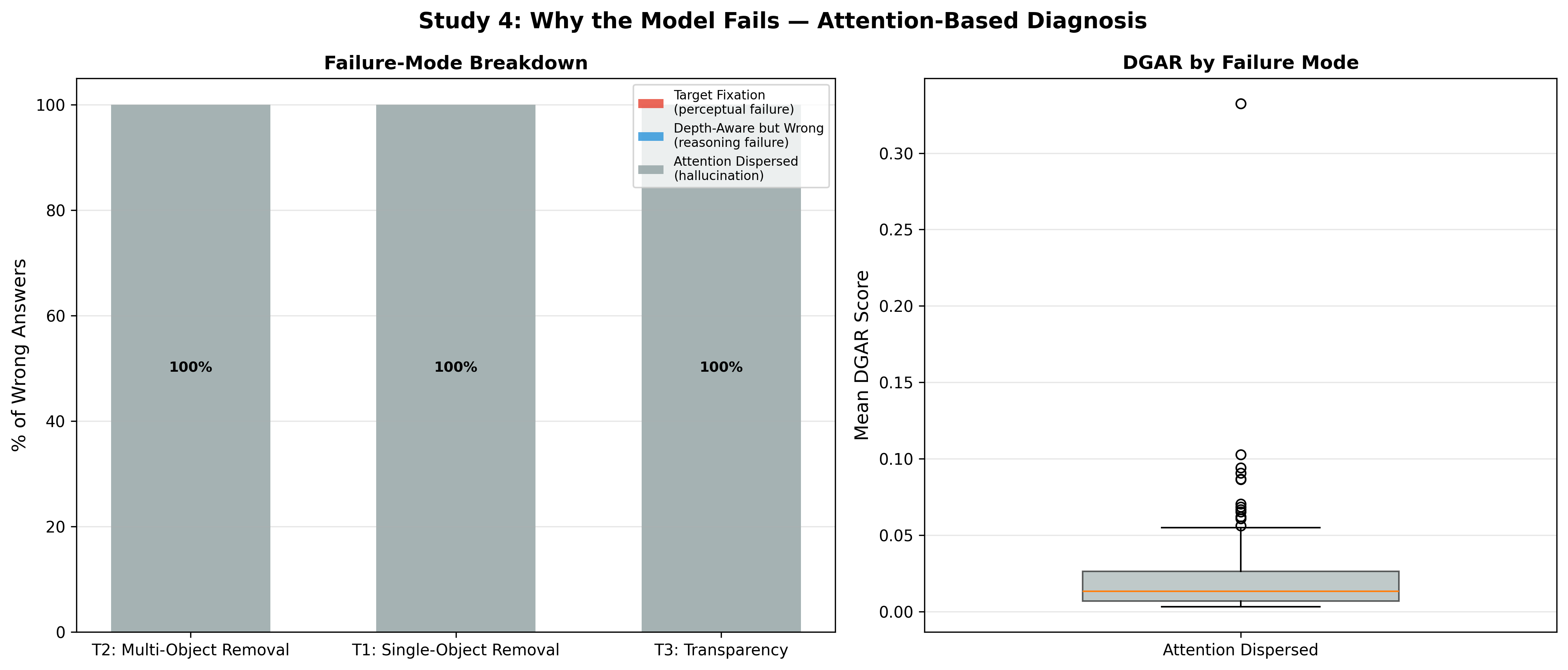}
  \caption{\textbf{DGAR aggregate findings over 163 failure cases.} \textit{Left:} mean attention fraction allocated to each region per task family; dashed line marks uniform chance ($1/3$). Across all three tasks, the depth-correct region receives well below 5\% of visual attention, while the irrelevant region absorbs roughly 80\%. \textit{Right:} failure-mode distribution. \emph{Attention Dispersed} accounts for 100\% of failures across all three tasks; no failure is explained by target fixation or by depth-aware-but-wrong reasoning.}
  \label{fig:dgar_main}
\end{figure}

\begin{figure}[t]
  \centering
  \includegraphics[width=0.49\textwidth]{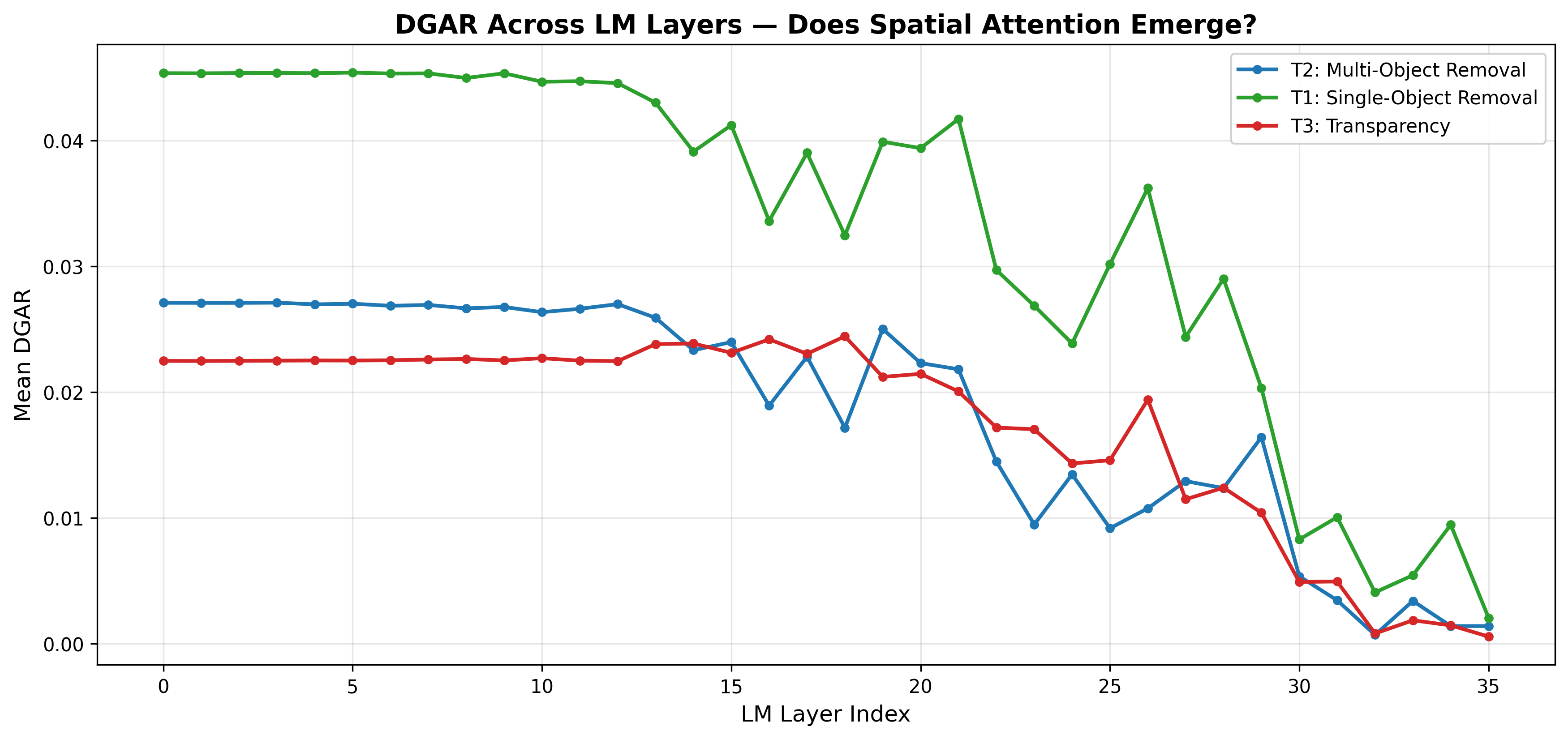}
  \hfill
  \includegraphics[width=0.49\textwidth]{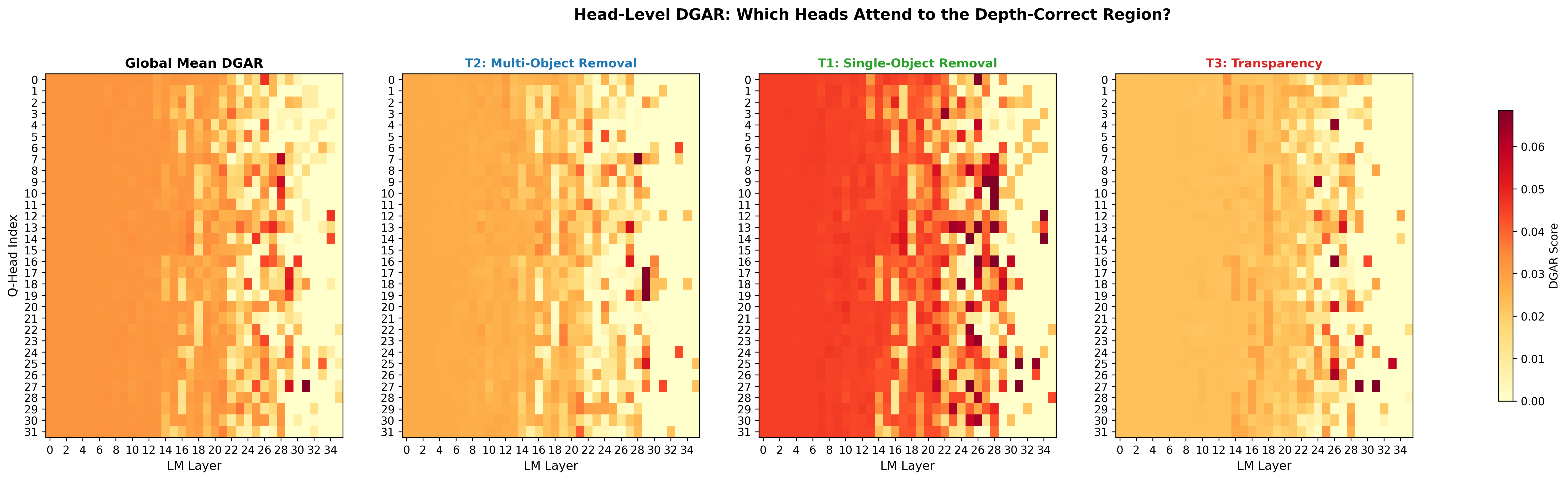}
  \caption{\textbf{DGAR is uniformly low across the LM decoder.}
  \textit{Left:} mean DGAR across the LM decoder layers; spatial
  attention to the depth-correct region remains consistently low across
  the entire decoder, drifting downward in deeper layers rather than
  rising. No layer shows a significant spike toward depth-correct
  attention. \textit{Right:} layer$\times$head DGAR heatmap; no
  individual head or layer emerges as a strong spatial specialist,
  consistent with the absence of dedicated depth-reasoning circuitry.}
  \label{fig:dgar_layer_head}
\end{figure}

\paragraph{Findings.}
Across all three counterfactual task families, the model allocates only a small fraction of its visual attention to the depth-correct region $\region{D}$ (well below the $1/3$ uniform-chance baseline), with $\region{I}$ absorbing roughly 80\% (\Cref{fig:dgar_main}, left). The failure-mode classification is unequivocal: \emph{Attention Dispersed} accounts for 100\% of failures across T1, T2, and T3 (\Cref{fig:dgar_main}, right); neither \emph{Target Fixation} nor \emph{Depth-Aware but Wrong} explains any of the 163 failures. Per-layer DGAR is uniformly low across the LM decoder and trends \emph{downward} in deeper layers; no head emerges as a spatial specialist (\Cref{fig:dgar_layer_head}). Attention to the depth-correct region is not lost in a particular layer; it never emerges in any layer.

\subsection{Causal Tracing via Activation Patching}
\label{sec:causal}

DGAR reveals \emph{where} attention falls but not whether it \emph{causally matters}. We complement it with activation patching~\cite{meng2022locating}, testing whether corrupting spatial representations changes the predicted token.

\paragraph{Setup.}
For each failure example we record $\prob_{\mathrm{clean}}(\tstar)$ for the top-1 token~$\tstar$, then run two corruptions: \textbf{A} (target-object patches) and \textbf{B} (depth-correct patches), using the DGAR spatial definitions mapped to the ViT patch grid. Each corrupted patch embedding entering the first ViT block is replaced with matched Gaussian noise. The probability drop $\probd = \prob_{\mathrm{clean}}(\tstar) - \prob_{\mathrm{corr}}(\tstar)$ classifies each example as \textsc{grounded} ($\probd > 0.05$ or argmax flip), \textsc{ungrounded} ($\probd < 0.01$ and argmax preserved), or \textsc{marginal} otherwise. For grounded examples, restoring layer~$\layervar$'s clean activation under corruption gives the recovery score
\begin{equation}
  \recovery_\layervar = \frac{\prob_{\mathrm{rest},\layervar}(\tstar) - \prob_{\mathrm{corr}}(\tstar)}{\prob_{\mathrm{clean}}(\tstar) - \prob_{\mathrm{corr}}(\tstar)}.
  \label{eq:recovery}
\end{equation}
We probe 17 sites: ViT blocks $\{0,6,12,18,23\}$, the main merger plus 3 deep-stack mergers, and LM layers $\{0,4,8,12,16,20,24,27\}$.

\begin{figure}[t]
  \centering
  \includegraphics[width=\linewidth]{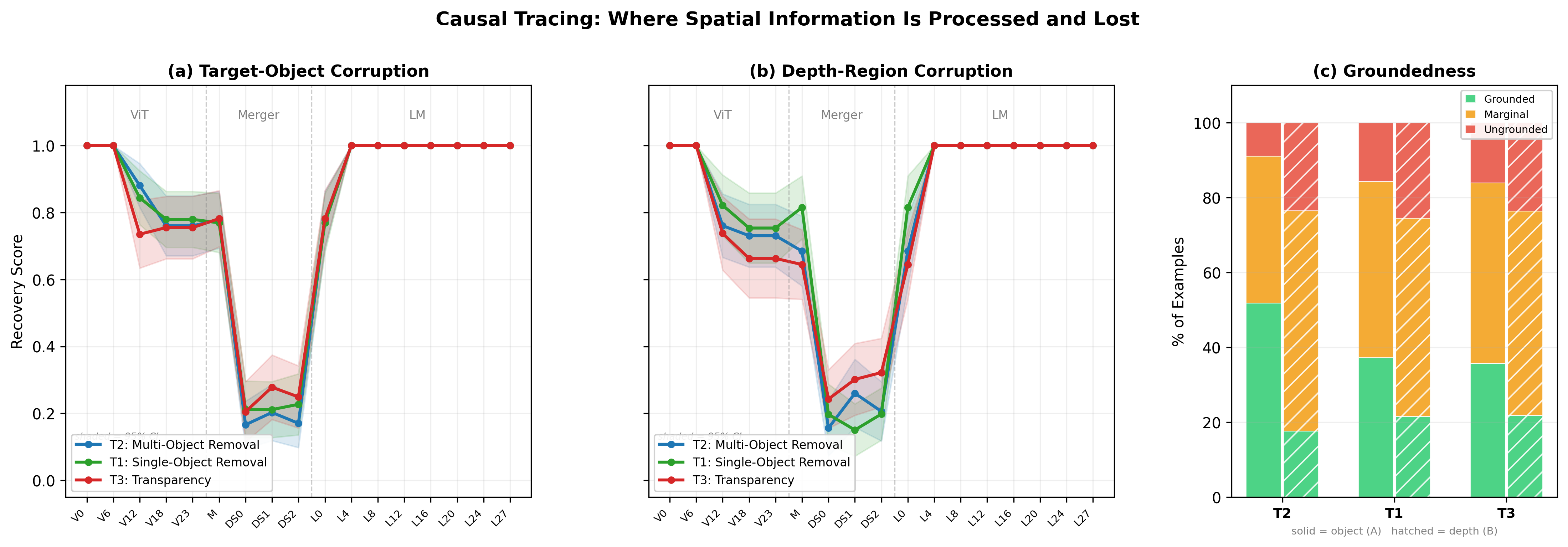}
\caption{\textbf{Causal tracing reveals where spatial information is processed and lost} (163 failure cases). (a)~Recovery curves for Corruption~A (target-object patches): restoration is ineffective at the merger stage but fully recovers at L0. (b)~Recovery curves for Corruption~B (depth-correct patches): identical V-shape with sharp drop at the merger. (c)~Groundedness classification per task and corruption type. Vertical dashed lines in (a, b) delimit ViT, merger, and LM stages; shaded bands in (a, b) denote 95\% confidence intervals over the 163 failure cases.}
  \label{fig:causal_tracing}
\end{figure}

\paragraph{The merger is the architectural bottleneck.}
\Cref{fig:causal_tracing}(a, b) reveals a striking signature across all 163 failure cases: recovery scores remain near 1.0 throughout the ViT blocks (V0--V23), collapse to 0.15--0.30 across the merger stage (M, DS0--DS2), and snap back to 1.0 at the first LM decoder layer (L0) for both corruption types. The merger pipeline is therefore the locus where spatial information becomes inaccessible: restoring activations \emph{after} the merger fully recovers the prediction, but restoring \emph{within} it does not. The V-shape signature is consistent across all three task families and both corruption types, and was first observed in a 37-case pilot before reproducing without qualitative change at the full 163-case scale.

\paragraph{Wrong answers depend on the named object more than on the depth-correct region.}
The groundedness analysis (\Cref{fig:causal_tracing}c) reveals a sharp asymmetry between corruptions. Under target-object corruption (A), the majority of failures are grounded or marginal; under depth-region corruption (B), the grounded fraction drops substantially and a larger share becomes ungrounded. The model's wrong answers are far less sensitive to the depth-correct region than to the named object itself, exactly the pattern DGAR predicts. Fully ungrounded cases (ungrounded under both corruptions) remain a small minority across all three tasks, indicating that wrong answers are typically driven by features of the named object rather than by pure language-prior hallucination.

\subsection{Cross-Study Synthesis}
\label{sec:cross}

The two studies converge on a consistent picture for \modelname{}: spatial structure is encoded inside the ViT but does not survive the merger pipeline (patching post-merger activations fully recovers the prediction; patching within does not), and at the language decoder, attention to the depth-correct region is uniformly dispersed. \emph{Attention Dispersed} explains 100\% of failures, with no spatial-specialist head at any layer. The asymmetry between target-object and depth-region corruptions further indicates a misgrounded simulation rather than language-prior fabrication. Extended four-part discussion is in Appendix~\ref{sec:appendix_mechanistic}. \emph{Scope.} These findings come from one open-weight VLM; replicating DGAR on Qwen3-VL-30B-A3B-Thinking is the most direct first test of generalisation.


\section{Discussion and conclusion}
\label{sec:discussion}

The empirical and mechanistic findings tell a coherent story across three levels. \emph{Behaviourally}, six VLMs spanning proprietary and open-weight systems and three orders of magnitude in active parameters succeed at volumetric planning over visible layouts (53--97\%, collision errors under 7\%) and fall to 6--45\% on depth-ordered occlusion (three independent probes converging) and 1--7\% on optical-geometry reasoning. \emph{Attentionally}, when \modelname{} fails on the three occlusion tasks, visual attention to the depth-correct region is well below uniform chance and the failure mode is uniformly \emph{Attention Dispersed} across all 163 cases we examined, with DGAR remaining low across every layer of the language decoder. \emph{Causally}, in the same model, the visual-token merger is where spatial structure becomes inaccessible to downstream computation: information is recoverable inside the ViT and, once restored, is recoverable again at the first LM layer.

\paragraph{What the deficit is not, and why this framing matters.} The deficit is not 2D-perceptual (object recognition is intact, and 3D collision constraints are computed correctly when the layout is visible), not a planning or chain-of-thought deficit (thinking-mode is enabled, and multi-step swaps are planned with high accuracy), and not a scaling deficit (the gap within the failure regime is dwarfed by the gap to the visible-state ceiling). It is structural: current VLMs capture \emph{what} a scene contains but not \emph{where} its parts sit in three-dimensional space. The dissociation reframes where current VLM limitations actually sit. On scenes with explicit 3D layout, frontier models reason and plan capably; the bottleneck is the structure the image does not display directly but the scene nonetheless contains, and that is what physical-world inference depends on.

\paragraph{Implications for VLM design.} The visual-token merger is commonly treated as a near-lossless compression step whose role is throughput rather than representation. Our causal trace points to a different reading: in the model we study, the geometric structure preserved through the merger is insufficient to support occlusion reasoning. Whether the merger plays the same role in the other five models (which share the ViT $\to$ token compression $\to$ language decoder pattern) is a question for follow-up work. Concrete architectural interventions the localization suggests include retaining geometric tokens through the merger, training an auxiliary depth-prediction objective at its output, or routing a subset of visual tokens directly to the language decoder.

\paragraph{Conclusion} VLMs see what is in the image. The 3D structure the image projects from, what lies behind a foreground object and what produces a reflection on a surface, is largely missing from the representations they pass to the language decoder. Our benchmark provides a clean, reproducible measurement of this gap across three competencies, and the mechanistic study points to a candidate architectural site whose redesign is worth probing in models beyond the one we instrumented.


\section{Limitations}
All scenes are synthetic (Nano Banana 2 \cite{raisinghani2026nanobanana}) and human-validated; a real-photograph replication of T1 to T4 is the most direct way to falsify a synthetic-artifact reading and is a planned extension. To the best of our knowledge, we were unable to find real world scene datasets suitable for this task. The mechanistic results come from one open-weight VLM; our scenes are single-viewpoint, tabletop-scale, indoor, and English-prompted; and the current 3{,}034-sample scale rests on end-to-end human annotation. Extended discussion of these limitations and a semi-automated construction pipeline for scaling are in Appendix~\ref{sec:appendix_scaling}.



\bibliographystyle{plainnat}
\bibliography{references}


\appendix

\begingroup
\hypersetup{linkcolor=black}
\renewcommand{\contentsname}{Appendix Contents}
\setcounter{tocdepth}{2}
{\small\tableofcontents}
\endgroup

\vspace{1em}
\noindent The appendix is organized as follows.
Appendix~\ref{sec:appendix_related_table} provides extended related-work discussion and a detailed comparison of our benchmark against prior benchmarks.
Appendix~\ref{sec:appendix_competencies_extended} gives the extended discussion of the three competencies and the design principles underlying the six tasks.
Appendix~\ref{sec:appendix_definitions} gives canonical visibility and occlusion definitions used to derive ground truths.
Appendix~\ref{sec:appendix_construction} details the construction pipeline (image generation, quality control, annotation).
Appendix~\ref{sec:appendix_prompts} lists the full evaluation prompt templates used for all six tasks.
Appendix~\ref{sec:appendix_impl} reports implementation details for the mechanistic analysis (model architecture, mask hyperparameters, compute, full DGAR equations).
Appendix~\ref{sec:appendix_mechanistic} contains additional mechanistic results, including the extended four-part cross-study synthesis.
Appendix~\ref{sec:appendix_reflection} presents a qualitative case study of a reflection (T4) failure.

\section{Extended Related Work and Detailed Benchmark Comparison}
\label{sec:appendix_related_table}

This appendix expands on the compact related-work treatment in \Cref{sec:related}.

\paragraph{Spatial reasoning in VLMs.}
Existing benchmarks evaluate spatial competencies over the \emph{currently visible} scene state. VSR~\cite{liu2023vsr}, SpatialRGPT~\cite{cheng2024spatialrgpt}, SpatialVLM~\cite{chen2024spatialvlm}, and InternSpatial~\cite{chen2025internspatial} test relation classification, region grounding, distance estimation, and large-scale spatial QA. Diagnostic suites~\cite{kamath2025mindgap,ranasinghe2024mindgap,tong2024vlmsblind,yang2024vsibench} target failure modes such as directional confusion. Embodied variants include Ego3D-Bench~\cite{chen2024ego3d}, NuScenes-SQA~\cite{qian2024nuscenesspatial}, and iVISPAR~\cite{lin2025ivispar}. Across this body of work, the model is asked about objects and relations the image already displays. None probes \emph{spatial counterfactual reasoning} or the optical-geometry inferences linking visible reflections to 3D producers.

\paragraph{Spatial mental modeling.}
MindCube~\cite{wang2025mindcube} assesses the capacity of VLMs to construct cognitive maps or mental models of a scene from limited, multi-view visual inputs, establishing a diagnostic framework for spatial inference that specifically tests whether models can maintain 3D consistency when moving between viewpoints (e.g., predicting what is visible from a new pose). MindCube primarily targets \emph{reconstruction} (mental mapping of observed views) and viewpoint transformation. Our benchmark differs by focusing on \emph{active intervention}: the ``what-if'' of removing, transparentising, or rearranging an object, which demands a deeper volumetric understanding of occlusion layers and depth ordering that limited-view reconstruction does not explicitly target.

\paragraph{Counterfactual reasoning.}
CounterVQA~\cite{mahmood2025countervqa} and COVER~\cite{wang2025cover} evaluate \emph{temporal} counterfactuals over video; ComPhy~\cite{chen2022comphy} and CausalVLBench~\cite{zhang2025causalvlbench} target \emph{causal} counterfactuals over physics simulations and structured causal models. Neither addresses the \emph{spatial} counterfactuals we introduce: static-scene interventions whose answers require an internal model of occlusion layers, depth ordering, and volumetric extent.

\paragraph{Interpretability.}
Activation patching and causal tracing have been used to localise factual recall in language models~\cite{meng2022locating}, attention heads in vision transformers~\cite{caron2021dino}, and modality-bridging behaviour in multimodal systems; we apply the same methodology to a behavioural deficit. We also eschew automated evaluation (LLM-as-Judge~\cite{zheng2024judging,li2023alpacaeval} and multimodal extensions~\cite{yu2024mmvet,liu2025mmbench}) in favor of end-to-end human evaluation, since 3D-structural questions require fine-grained visual inspection that cannot be reliably delegated to a model exhibiting similar deficits.

\Cref{tab:related_comparison} compares our benchmark against representative prior benchmarks along the dimensions most relevant to this work.

\begin{table}[h]
\centering
\caption{\textbf{Comparison with related benchmarks.} Spatial = spatial relations over visible state; CF = counterfactual reasoning (any kind: causal, temporal, or spatial); Occl.\ = occlusion / visibility under intervention; Rearr.\ = rearrangement planning. Our benchmark is the only entry that combines spatial counterfactuals with occlusion-under-intervention probing under human evaluation.}
\label{tab:related_comparison}
\resizebox{\linewidth}{!}{%
\begin{tabular}{lcccccl}
\toprule
\textbf{Benchmark} & \textbf{Spatial} & \textbf{CF} & \textbf{Occl.} &
\textbf{Rearr.} & \textbf{Eval.} & \textbf{Domain} \\
\midrule
VSR~\cite{liu2023vsr}              & \cmark & \xmark & \xmark & \xmark & Auto & Natural images \\
SpatialRGPT~\cite{cheng2024spatialrgpt} & \cmark & \xmark & \xmark & \xmark & Auto & Natural images \\
InternSpatial~\cite{chen2025internspatial} & \cmark & \xmark & \xmark & \xmark & Auto & Multi-view \\
Mind the Gap~\cite{kamath2025mindgap} & \cmark & \xmark & \xmark & \xmark & Auto & Diagnostic pairs \\
iVISPAR~\cite{lin2025ivispar}      & \cmark & \xmark & \xmark & \cmark & Auto & Tile puzzles \\
CounterVQA~\cite{mahmood2025countervqa} & \xmark & \cmark & \xmark & \xmark & Auto & Video \\
COVER~\cite{wang2025cover}          & \xmark & \cmark & \xmark & \xmark & Auto & Video \\
ComPhy~\cite{chen2022comphy}        & \xmark & \cmark & \xmark & \xmark & Auto & Physics sim. \\
CausalVLBench~\cite{zhang2025causalvlbench} & \xmark & \cmark & \xmark & \xmark & Auto & Mixed \\
\midrule
\textbf{Ours} & \cmark & \cmark & \cmark & \cmark & \textbf{Human} & \textbf{Synthetic (validated)} \\
\bottomrule
\end{tabular}%
}
\end{table}

\section{Extended Discussion: The Three Competencies and the Six Tasks}
\label{sec:appendix_competencies_extended}

This appendix expands the compact taxonomy of \Cref{sec:benchmark} with the full
motivation for each competency and the design principles underlying the six tasks.

\paragraph{Competency 1: Depth-ordered occlusion via spatial counterfactuals (T1, T2, T3).}
A 3D scene representation encodes which objects sit in front of which. A flat 2D catalogue does not. We probe this competency through spatial counterfactuals: interventions on the scene whose answers are not present in the image and can only be derived from a depth-ordered model of it. We use three independent operationalisations so that any consistent failure pattern reflects the underlying capability rather than a single prompt formulation.

\emph{T1 (single-object removal)} asks: ``if X is removed, what becomes visible?'' This is the simplest probe and isolates pairwise depth ordering: to answer correctly, the model must know which object sits directly behind X.

\emph{T2 (multi-object removal)} asks: ``what is the smallest set of objects that must be removed so X becomes fully visible?'' This adds combinatorial inference: the model must reason about chains of occlusion, identifying every object that contributes to occluding X and selecting the minimum cover.

\emph{T3 (transparency)} asks: ``if X were transparent, what would be visible through it?'' This adds volumetric depth reasoning: the model must treat X as a 3D volume with extent, not a 2D silhouette, and identify what lies along rays passing through that volume.

The three tasks deliberately differ in surface phrasing, in the direction of the counterfactual (deletion vs.\ transparency), and in the type of inference required (depth ordering, set selection, ray-casting). If the deficit lies in occlusion reasoning itself, in the absence of a depth-ordered scene representation, the three should fail together. If the deficit is task-specific, they should not.

\paragraph{Competency 2: Optical geometry (T4).}
A 3D scene representation encodes optical projection relationships: it knows that a reflection on a surface comes from a specific 3D object whose geometry produces that reflection under the scene's lighting and viewpoint. A flat 2D catalogue can detect that a reflective surface exists but cannot trace which object produces which reflection. T4 isolates this capability: ``which objects have their reflection visible in any surface?'' Crucially, the reflections themselves \emph{are} present in the image, so this is not a hidden-state question. The failure mode tested here is purely about cross-region geometric reasoning: linking a 2D reflection patch on a surface to the 3D object that projects onto it.

\paragraph{Competency 3: Volumetric planning (T5, T6).}
A 3D scene representation encodes object volumes and their physical occupancy of space. We test this through rearrangement planning under collision constraints. \emph{T5 (single swap)} asks the model to choose one pairwise swap that achieves a target left-to-right ordering without causing overlaps; \emph{T6 (multi-swap)} extends this to multi-step search. T5 and T6 differ from T1--T4 in a critical respect: the 3D structure needed to answer the question is essentially \emph{given} by the layout in the image. The objects are visible, their 2D footprints are visible, and the model needs to compute whether two of them would collide if swapped, a question about volumes the model can already see. This is the most permissive 3D test in the benchmark and serves as a diagnostic floor: if a model fails Volumetric planning, we know its 3D failure is not the depth-and-projection-inference failures that T1--T4 test, but something more fundamental.

\paragraph{Why these six together.} The taxonomy is not arbitrary. T1--T3 force the model to \emph{infer} 3D structure that is hidden behind foreground objects; T4 forces it to \emph{trace} 3D structure across optical projection; T5--T6 ask whether it can \emph{use} 3D structure that is already visible. Each task tests a competency that physical-world deployment will require, and no task is reducible to another. T1, T2, T3 fail together (they probe the same underlying representation). T4 fails for an independent reason (it requires cross-region projection, not depth inference). T5/T6 reveal what \emph{works} in the current paradigm, isolating the failure modes from the successes.

\paragraph{Design principles.}
\textbf{(i)~Non-trivial 3D structure.} Every task is constructed so that a flat object-catalogue representation cannot produce the right answer.
\textbf{(ii)~Triangulated probing.} Occlusion reasoning is probed three independent ways (T1, T2, T3) so that any consistent failure pattern reflects the underlying competency.
\textbf{(iii)~Diagnostic spread.} The six tasks span the three competencies, with at least one task probing each, so failures and successes both carry information.
\textbf{(iv)~Deterministic ground truth.} Every question admits a unique answer under canonical visibility and occlusion definitions.
\textbf{(v)~Three-way scoring with no partial credit.} Each response is graded as Correct, Incorrect, or Hallucinated; partial-match answers are graded Incorrect.

\paragraph{Hallucination definition.}
We define \emph{hallucination} as any case where the model names an object that is clearly inconsistent with the scene, including both objects absent from the image and misidentifications substantial enough that a human annotator judges the label incorrect. This inclusive operationalisation follows established practice in object-hallucination evaluation~\cite{rohrbach2018object,wang2023amber}, where lexical mismatches against ground-truth labels are treated as hallucinations alongside absent-object references, both reflecting reliance on language priors over visual evidence.

\paragraph{Natural-language prompts.}
All prompts are written in \emph{natural language}, referring to objects by everyday noun phrases (e.g.\ ``\texttt{flower pot}'', ``\texttt{double tap beer tower}''). We deliberately do \emph{not} provide bounding boxes, segmentation masks, point coordinates, numeric IDs, or any other explicit spatial pointer to the target. The capability we are measuring is whether a VLM can \emph{itself} ground a linguistic reference to the correct image region and then reason about its 3D context. Supplying coordinates would short-circuit precisely the grounding step we are testing, reducing the task to retrieval over a pre-localised region. Natural-language prompting also matches the conditions under which VLMs are deployed downstream (in robotic instruction following, embodied question answering, and multimodal assistance) where users describe objects in words rather than pixels. The benchmark therefore measures the end-to-end capability that matters for real use, not a coordinate-fed upper bound on it. We additionally ensure that natural-language references resolve unambiguously within each scene: the image-generation pipeline (Appendix~\ref{sec:appendix_imggen}) enforces that no two sampled objects share the same base-type noun, so a noun phrase like ``\texttt{flower pot}'' has a unique referent in the scene by construction. In the small number of cases where the generated image nonetheless contains visually similar duplicates that survive this constraint, the duplicated objects are excluded from both the prompt target and the ground-truth set for every task in that scene; an ambiguous referent is never asked about, and never required as an answer.

\paragraph{Limitations.}
All images are single-viewpoint, tabletop-scale indoor scenes. The benchmark characterises specific deficits in occlusion, optical-geometry, and volumetric reasoning but does not claim exhaustive coverage of 3D scene understanding more broadly. Synthetic images can in principle carry distributional artifacts (lighting, shadow consistency, material rendering) that affect VLMs trained primarily on natural images for reasons unrelated to 3D structure. We mitigate this by validating every image. A synthetic-artifact reading would also have to explain why the same models, evaluated on images generated by the same pipeline, succeed at T5/T6 (rearrangement over visible state) while collapsing on T1--T4 (counterfactual simulation): if the failures were driven by rendering artifacts, the artifacts would have to selectively impair occlusion reasoning while leaving object recognition, collision judgement, and rearrangement planning intact. A real-photograph replication of T1 and T4 is the most direct way to falsify a synthetic-artifact reading and is a planned extension.

\section{Canonical Visibility and Occlusion Definitions}
\label{sec:appendix_definitions}

This section will contain the canonical definitions used by annotators
to derive deterministic ground truths: \emph{full visibility} (no
occluding object intersects the silhouette of the target from the
camera viewpoint), \emph{partial occlusion} (any silhouette overlap),
and the per-task projection rules (single-object removal, multi-object
removal, volumetric transparency, structural reflection). The three-way
scoring rubric (Correct, Incorrect, Hallucinated) and tie-breaking
conventions used by QC reviewers are also documented here.

\section{Construction Pipeline}
\label{sec:appendix_construction}

\subsection{Image Generation Pipeline}
\label{sec:appendix_imggen}

All scenes are synthetically generated to control occlusion structure,
object diversity, and visual fidelity to real indoor environments. The
pipeline uses two distinct routines: one for reflection scenes (T4)
and one for occlusion-based scenes (T1--T3, T5--T6).

\paragraph{Occlusion scenes (T1, T2, T3, T5, T6).}
For each scene, we sample one of 15 indoor scene templates (kitchen
counter, dining table, office desk, hallway console, mudroom bench,
craft room, etc.) and 50 objects from a curated pool of 1{,}800+
medium-to-large everyday items, ensuring no two sampled objects share
the same base-type noun (so e.g.\ ``duffel bag'' and ``messenger bag''
never co-occur). Objects are partitioned into occlusion chains of 3--5
items, with explicit per-position depth targets (50\% frame height at
1\,m, 25\% at 2\,m, 15\% at 3\,m, 8\% at 4\,m, 5\% at 5\,m), lateral
positions across nine slots, and a strict bare-walls constraint to
prevent shortcut exploitation. Images are generated with
\textbf{Nano Banana 2 ~\cite{raisinghani2026nanobanana}} at 3:2 aspect ratio, 1K resolution, then
resized to $720{\times}480$. We use a fresh random seed per image.

\paragraph{Reflection scenes (T4).}
Each reflection scene contains 7 objects placed on a reflective surface
sampled from 20 templates spanning polished wood, kitchen counters
(granite, marble, quartz), and glass. The base image is generated with
Nano Banana 2 showing all 7 objects with visible reflections.
Annotators then select 2 of the 7 objects whose reflections must be
removed, and the image is edited in \textbf{Adobe Photoshop} to remove
those specific reflections while preserving the remaining 5. The final
T4 ground truth is therefore the 5 objects whose reflections remain
visible after editing. This deliberate partial removal is a stress test: a model that infers reflections from a generic ``shiny surface implies reflections'' prior, rather than from the actual visible reflection patches, will list all 7 objects and fail. Correctly identifying only the 5 surviving reflections requires localising and tracing each reflection patch back to its 3D producer, the precise capability T4 is designed to probe.

\paragraph{Quality control on generated images.}
Annotators visually verified every generated image for: (i)~adherence
to the depth and occlusion structure required by the prompt;
(ii)~clutter and realism characteristic of natural indoor scenes;
(iii)~no text, logos, or branded surfaces (which would offer trivial
language shortcuts). Images failing any check were regenerated. All
final samples are human-validated.

\subsection{Annotation Protocol}
\label{sec:appendix_annotation}

All ground-truth answers and model-response evaluations are produced by
trained human annotators. The team consists of \textbf{15 annotators}
and \textbf{5 QC reviewers}, where the QC reviewers are former
top-performing annotators promoted based on accuracy and task
understanding. Every sample passes through two stages.

\paragraph{Stage 1: Annotation.}
For each image, an annotator (i)~validates the generated image (clutter,
occlusion structure, no-logo constraint), (ii)~writes the
task-specific prompt by filling the template with the appropriate
\texttt{<object>} or object list, (iii)~constructs the deterministic
ground-truth answer set, and (iv)~queries all six models and scores each
of their responses. End-to-end time per sample is approximately
\textbf{30 minutes}.

\paragraph{Stage 2: QC review.}
Every sample is reviewed by a QC reviewer who independently verifies
the prompt, the ground truth, and all six model scores. QC workload is
distributed across the 5 reviewers based on availability; \emph{every
sample receives QC review}. Disagreements are resolved by re-examining
the image and applying the canonical visibility/occlusion definitions
in Appendix~\ref{sec:appendix_definitions}.

\paragraph{Scoring rubric.}
Responses are scored on a strict three-way scale with no partial credit:
\begin{itemize}[nosep,leftmargin=1.5em]
  \item \textbf{Correct:} the model's answer set exactly matches the
    ground-truth set. For T2 the order of removed objects must also
    match; for all other tasks the set comparison is unordered. Naming
    2 of 3 ground-truth objects is graded \emph{Incorrect}, not partial
    credit; a near-miss is treated identically to any other wrong
    answer.
  \item \textbf{Incorrect:} the answer is wrong but every named object
    is present in the scene. The model misidentifies the relationships
    in the scene without fabricating new objects. Incorrect responses
    are full failures: they receive no credit and contribute equally to
    the error count regardless of how close they are to the ground
    truth.
  \item \textbf{Hallucinated:} the model's response references one or
    more entities that the annotator cannot discernibly identify in the
    scene image, or substantially misidentifies an object relative to
    its canonical label. Hallucination is determined by the annotator
    visually verifying every named object against the image, in line
    with the inclusive definition adopted in \Cref{sec:benchmark}.
    Hallucinated responses are also full failures and are reported
    separately from Incorrect responses to expose the language-prior
    failure mode.
\end{itemize}
Reported accuracy in the main paper is the fraction of responses graded
\emph{Correct}. Both \emph{Incorrect} and \emph{Hallucinated} responses
count against the model.

\section{Evaluation Prompt Templates}
\label{sec:appendix_prompts}

Every model receives one of six fixed prompt templates corresponding to
its task. We use canonical templates with no chain-of-thought hints to
isolate the model's intrinsic reasoning. As discussed in
\Cref{sec:benchmark}, all prompts refer to objects by natural-language
noun phrases and provide \emph{no} bounding boxes, masks, point
coordinates, or numeric IDs; the model must perform the
linguistic-to-spatial grounding step itself.

\begin{itemize}[nosep,leftmargin=1.5em]
  \item \textbf{T1 (Single-Object Removal):} ``If \texttt{<object>} is removed, which
    previously partially occluded objects become fully visible, from
    the current fixed camera viewpoint? Give me only the list of
    objects and no additional commentary or information.''
  \item \textbf{T2 (Multi-Object Removal):} ``Which object instance(s)
    comprise the minimum set that must be removed so that
    \texttt{<object>} becomes fully visible from the current fixed
    camera viewpoint? Give me only the list of objects and no
    additional commentary or information.''
  \item \textbf{T3 (Transparency):} ``If \texttt{<object>}, all its
    internal components, and anything contained within it are made
    perfectly transparent, which objects are now visible through its
    volume, from the current fixed camera viewpoint? Give me only the
    list of objects and no additional commentary or information.''
  \item \textbf{T4 (Reflection):} ``Which objects have their
    reflection visible in any surface, from the current fixed camera
    viewpoint? Give me only the list of objects and no additional
    commentary or information.''
  \item \textbf{T5 (1-Swap):} ``Which pair of objects should I swap
    (one pairwise swap allowed), without moving any other object and
    without causing any overlaps or collisions, so that the relative
    left-to-right order by x-coordinate alone (y-coordinates are
    irrelevant) of \texttt{<obj1>}, \texttt{<obj2>}, \texttt{<obj3>},
    \texttt{<obj4>} matches the listed order, from the current fixed
    camera viewpoint? Give me only the pair of objects and no
    additional commentary or information.''
  \item \textbf{T6 (Multi-Swap):} same as T5 but with ``one or more
    pairwise swaps allowed''.
\end{itemize}

All six models are queried via the OpenRouter API at \textbf{temperature
0}, single inference pass per sample, no system prompt beyond the task
prompt itself, and \textbf{thinking mode enabled} via the standard
\texttt{enable\_thinking} flag (all six models support it). We do not
use few-shot examples, retrieval, or any auxiliary tools.

\section{Mechanistic Implementation Details}
\label{sec:appendix_impl}

\paragraph{Model.}
We study \modelname{}, a vision--language model with a 27-block ViT
(patch size~16, dim~1152, 16 attention heads, head dim~72; deepstack
taps at blocks 8, 16, 24), a $2{\times}2$ spatial merger ($4096 \to 4096$
dim) plus three deep-stack mergers, and a 36-layer LM decoder
(dim~4096, 32 Q-heads, 8 KV-heads with GQA, head dim~128).

\paragraph{Depth estimation.}
Monocular depth maps are computed with Depth-Anything-V3
Large~\cite{depthanything3}. Confidence maps are synthesised from gradient magnitude;
a threshold of 0.5 excludes unreliable regions.

\paragraph{Object masks.}
Per-object masks are generated using SAM~3~\cite{carion2025sam3segmentconcepts} with text
prompts derived from the task query. Bounding box expansion fractions are
task-specific: 0.25 (T2, multi-object removal), 0.375 (T1, single-object
removal), and 0.125 (T3, transparency). Expansion is applied
per-side, so total bounding box growth is twice the listed fraction.
The largest expansion (T1) accommodates occluded objects whose extent
can exceed the occluder's footprint, while the smallest (T3) reflects
the fact that objects revealed through a transparent target lie
predominantly directly behind it. Masks are not distributed; users of
the released mechanistic code generate them per image for the target
object using SAM~3.

\paragraph{Full DGAR equations.}
For completeness we restate the full DGAR formulation. Let
$\mathbf{M} \in \{0,1\}^{H \times W}$ be the binary object mask and
$\mathbf{D} \in \mathbb{R}^{H \times W}$ the monocular depth map. The
target depth is
\[
  d_{\mathrm{obj}}
  = \mathrm{median}\bigl(\mathbf{D}[\mathbf{M} > 0]\bigr).
\]
For T2 (multi-object removal) the depth-correct condition is
$\mathbf{D} < d_{\mathrm{obj}} - \delta$ (shallower than the target;
selects occluders in front), while for T1 (single-object removal)
and T3 (transparency) it is
$\mathbf{D} > d_{\mathrm{obj}} + \delta$ (deeper than the target;
selects what is hidden behind), where $\delta$ is a depth margin. We set
$\delta = 0.3$\,m and fall back to an adaptive margin of
$0.1 \times (\max - \min)$ of the local depth range when the fixed
margin yields an empty region.

For layer~$\layervar$ and head~$\headvar$, the attention weight over
visual token~$i$ is
\begin{equation}
  \alpha_i^{(\layervar,\headvar)}
  = \frac{
    \exp\!\left(\mathbf{q}^{\top} \mathbf{k}_i / \sqrt{d}\right)
  }{
    \sum_{j=1}^{V} \exp\!\left(\mathbf{q}^{\top} \mathbf{k}_j / \sqrt{d}\right)
  }.
  \label{eq:attn_app}
\end{equation}

\paragraph{Computational cost.}
DGAR requires one forward pass per example. Causal tracing requires
$1 + C + 17C$ passes ($C \in \{1,2\}$ corruption types). 

\section{Additional Mechanistic Results}
\label{sec:appendix_mechanistic}

\paragraph{Extended four-part diagnosis.}
\Cref{sec:cross} summarises the synthesis of DGAR and activation
patching in a compact form. The full four-part diagnosis, made
statistically clean by the 163-case sample, is as follows.
\textbf{(i)~Attention is dispersed and well below chance for the
depth-correct region}, with \emph{Attention Dispersed} accounting for
100\% of failures across all three task families and no
spatial-specialist head emerging at any LM layer.
\textbf{(ii)~The merger pipeline is where spatial information becomes
inaccessible to downstream computation.} Causal tracing localises the
transition unambiguously: spatial structure is encoded in the ViT but
is no longer recoverable from compressed tokens, while patching a clean
post-merger activation back in fully restores the prediction. By the
time the LM sees visual tokens, they no longer carry the depth structure
occlusion reasoning requires, consistent with the uniformly low DGAR
over LM layers.
\textbf{(iii)~Wrong answers depend on the named object more than on the
depth-correct region}: the model \emph{looks at} the named object and
produces an answer, but rarely consults the spatial neighbourhood where
the answer lives, consistent with the dispersed-attention finding from
DGAR.
\textbf{(iv)~Pure hallucination is rare; misgrounded simulation is the
norm.} Only a small minority of failures are fully ungrounded under
both corruptions. The deficit in this model is not primarily a
language-prior hallucination problem; it is a misgrounded simulation
problem in which the model uses the wrong visual evidence.

\paragraph{Effective sample counts for causal tracing.}
The mechanistic study covers 163 failure cases drawn from the three
counterfactual task families: $n=56$ for T1 (single-object removal),
$n=51$ for T2 (multi-object removal), and $n=56$ for T3 (transparency).
Under Corruption~A (target-object patches) all 163 examples are
analysed. Under Corruption~B (depth-correct region patches), 5 examples
from T2 are excluded because no patches satisfied
$\mathbf{D} < d_{\mathrm{obj}} - \delta$ (i.e.\ no shallower-than-target
region existed within the expanded bounding box, even with the adaptive
depth-margin fallback), giving $n_B = 51$ for T2; T1 and T3 retain the
full count under Corruption~B. The within-task patterns reported in
\Cref{sec:causal} (V-shape recovery curve, asymmetry between Corruption
A and B) are stable across all three task families at this sample size.

\paragraph{Detailed DGAR qualitative grid.}
We extend the aggregate DGAR analysis to the full set of failure
examples in the dataset using a per-example visualisation grid. The
visualisation grid uses a 4-column layout per example:
\textbf{(i)~Original image} with the target object outlined in red;
\textbf{(ii)~DA3 depth map} colorized with the \texttt{viridis} colormap
(near $\to$ warm, far $\to$ cool);
\textbf{(iii)~Depth-correct region} overlay (semi-transparent green
over patches satisfying the prompt-type-specific depth condition);
\textbf{(iv)~Model attention heatmap} averaged across the top-5 DGAR
heads at the peak DGAR layer for that example.

Examples are grouped by the three failure modes from the DGAR taxonomy:
\begin{itemize}[nosep,leftmargin=1.5em]
  \item \emph{Attention Dispersed:} attention scattered across irrelevant
    patches with no concentration on either the target or the
    depth-correct region.
  \item \emph{Target Fixation:} attention concentrated on the named
    object itself, ignoring the depth-correct region.
  \item \emph{Depth-Aware but Wrong:} attention does land on the
    depth-correct region, yet the model still produces an incorrect
    answer, indicating a downstream reasoning failure rather than a
    grounding failure.
\end{itemize}
At the 163-case sample size, every failure we examined falls into the
\emph{Attention Dispersed} category, and the qualitative grid documents
this uniformity. We additionally include success cases (correct
answers) where attention concentrates on the depth-correct region,
demonstrating that DGAR distinguishes meaningful from non-meaningful
attention allocation.

\section{Qualitative Case Study: Reflection Failure (T4)}
\label{sec:appendix_reflection}

To complement the aggregate findings, we trace a single forward pass
through the model's vision encoder and inspect spatial clustering at
the vision--language bridge. We focus on late vision blocks V21--V27,
which encode high-level visual structure, and on the bridge layer L20
of the language model, where visual evidence is integrated with the
linguistic query.

\begin{figure}[t]
\centering
\setlength{\fboxrule}{0.5pt}\setlength{\fboxsep}{4pt}
\begin{minipage}[c]{0.42\linewidth}
\centering
\includegraphics[width=\linewidth]{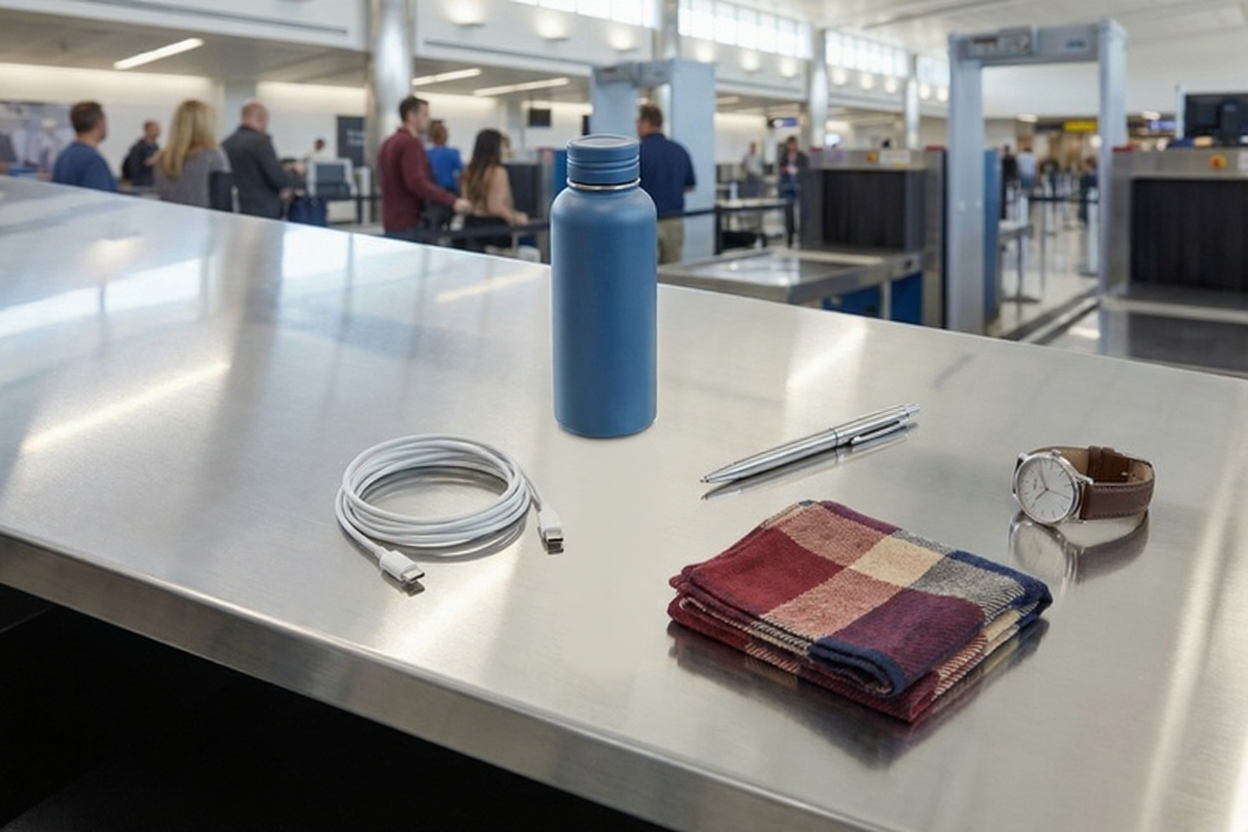}
\end{minipage}\hfill
\begin{minipage}[c]{0.55\linewidth}
\fbox{\parbox{\dimexpr\linewidth-10pt\relax}{\footnotesize\raggedright \textbf{Prompt:} ``Which objects have their reflection visible on any surface in this image?''}}\\[4pt]
\fbox{\parbox{\dimexpr\linewidth-10pt\relax}{\footnotesize\raggedright \textbf{Model response:} \textit{To determine which objects have visible reflections, we analyze the metallic table (a reflective surface). Any object placed on this surface will cast a reflection. The objects on the table are the blue water bottle, white USB cable, silver pen, brown leather watch, and checkered cloth; all these items rest on the reflective metallic table, so their reflections are visible on the table's surface.}}}
\end{minipage}
\caption{\textbf{Reflection task setup.} Scene (left); prompt and model response (right). The model lists \emph{every} object on the metallic countertop as having a visible reflection, including the blue water bottle and pen whose reflections are not resolved in the image. The reasoning chain reveals the failure: the model invokes the rule ``the table is reflective, so any object placed on it must have a reflection,'' applying a learned prior rather than verifying which reflections are visually present.}
\label{fig:reflection_setup}
\end{figure}

\paragraph{Setup.}
We probe the model with the prompt ``Which objects have their reflection
visible on any surface in this image?'' on an airport-counter scene
containing a blue water bottle, a coiled USB cable, a pen, a watch, and
a checkered cloth resting on a metallic countertop. A correct answer
would list only objects whose reflections are actually present on the
metal surface (in this scene, primarily the checkered cloth and the
watch, whose reflections are clearly resolved). \Cref{fig:reflection_setup}
shows the input and the model's response.

\begin{figure}[t]
\centering
\includegraphics[width=\linewidth,height=0.26\textheight,keepaspectratio]{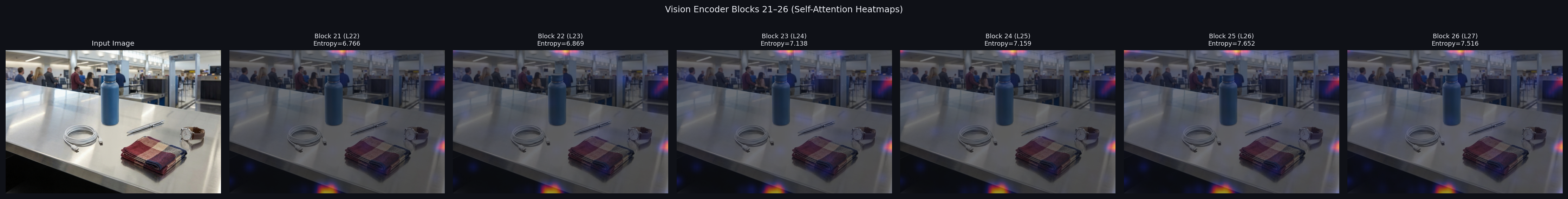}
\caption{\textbf{Vision encoder activation (V21--V27).} Self-attention concentrates on the foreground objects (bottle, cloth, watch) rather than on the countertop region containing the actual reflections. Entropy values are reported above each block.}
\label{fig:reflection_vision_encoder}
\end{figure}

\begin{figure}[t]
\centering
\includegraphics[width=\linewidth,height=0.26\textheight,keepaspectratio]{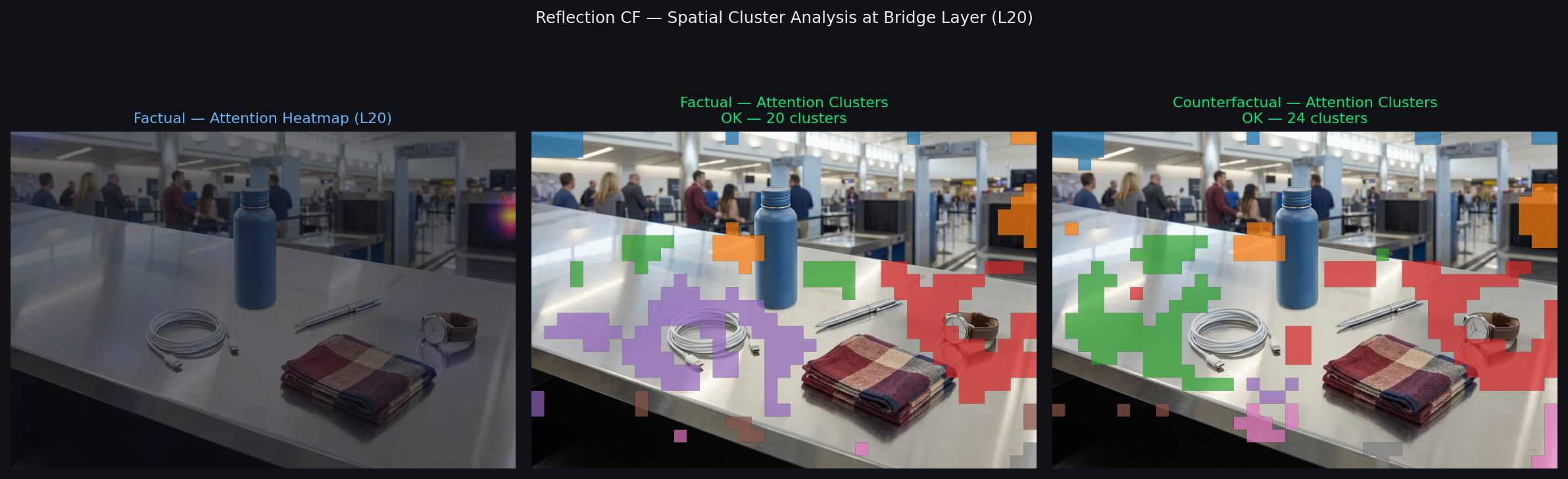}
\caption{\textbf{Bridge-layer cluster map (L20).} Left: factual attention heatmap. Centre: 20 attention clusters under the factual prompt. Right: 24 clusters under the counterfactual prompt. Both partitions are diffuse and lack the coherent object--reflection pair structure that a reflection-aware model would exhibit.}
\label{fig:reflection_cluster_map}
\end{figure}

\paragraph{Mechanism.}
Two observations identify the failure as a \emph{spatial--semantic
grounding} error rather than a perceptual one. First, late vision
blocks V21--V27 place their strongest activation on the foreground
objects themselves (bottle, cloth, watch) rather than on the
countertop region where reflections would appear
(\Cref{fig:reflection_vision_encoder}). The encoder sees the objects
clearly but never localises the reflection patches as a separate
visual entity. Second, at the vision--language bridge (L20), spatial
clustering is diffuse: 20 disjoint attention clusters under the
factual prompt and 24 under the counterfactual prompt
(\Cref{fig:reflection_cluster_map}), with no coherent
object--reflection \emph{pair} structure that would link an object to
its reflection on the surface.

Together, these signatures show that the model's response is generated
from a generic \emph{shiny-surface-implies-reflections} prior rather
than from visual evidence: the listed objects are exactly the objects
on the table, regardless of whether each one's reflection is actually
visible.

\newpage

\section{Scaling the Benchmark: A Semi-Automated Construction Pipeline}
\label{sec:appendix_scaling}

The current release is built end-to-end by trained human annotators (Appendix~\ref{sec:appendix_annotation}), at roughly 30 minutes per sample including QC. This was a deliberate choice: 3D-structural ground truth is precisely the kind of judgement we would not trust an LLM-as-judge to produce, because the models being judged exhibit the very deficits the benchmark targets. The price is scale: the 3{,}034-sample release exhausts what is reasonable to support with this protocol, and growing the benchmark to the $10^4$--$10^5$ regime needed to slice results by scene type, occluder count, or depth gradient is not feasible by human labour alone.

This appendix outlines the construction pipeline we plan to use to scale the benchmark, organised around two principles: \textbf{(i)~ground truth is derived from geometry-grounded perception modules, not from a VLM judging another VLM}, and \textbf{(ii)~the pipeline is validated by full two-layer human review of an initial 3{,}034-sample pilot batch before any further scale-out} (rather than by per-module agreement against the existing release) because every scene admits many valid configurations and a per-sample agreement target is the wrong objective. Once the pipeline produces a 3{,}034-sample batch that clears two-layer human review with no outstanding issues, it is treated as reliable and scaled to arbitrary size.

\subsection{Per-Task Construction Modules}
\label{sec:appendix_scaling_modules}

Each task admits a different decomposition into geometry-grounded operations. We describe each in turn.

\paragraph{Occlusion family (T1, T2, T3).}
The three occlusion tasks share a common spatial-grounding pipeline:
\begin{enumerate}[leftmargin=1.5em,nosep]
  \item \textbf{Object inventory.} Given a scene image, run an open-vocabulary detector and SAM~3~\cite{carion2025sam3segmentconcepts} to obtain per-object masks $\{M_i\}$ and canonical noun-phrase labels $\{\ell_i\}$. Reject scenes where any object label has confidence below a calibrated threshold.
  \item \textbf{Depth assignment.} Run Depth-Anything-V3~\cite{depthanything3} to obtain a per-pixel depth map $\mathbf{D}$. Assign each object a representative depth $d_i = \mathrm{median}(\mathbf{D}[M_i > 0])$.
  \item \textbf{Occlusion-graph construction.} For each ordered pair $(i, j)$ with $d_i < d_j$ and $\mathrm{IoU}(M_i, \mathrm{bbox}(M_j)) > \tau_{\mathrm{occ}}$, declare $i$ \emph{occludes} $j$. The resulting directed graph $G_{\mathrm{occ}}$ encodes the depth-ordered occlusion structure of the scene.
  \item \textbf{Per-task ground truth derivation.}
    \emph{T1 (single-object removal):} For target $X$, the ground truth is the set of $j$ such that $X$ is the \emph{unique} predecessor of $j$ in $G_{\mathrm{occ}}$ (i.e.\ removing $X$ leaves $j$ unoccluded).
    \emph{T2 (multi-object removal):} Compute the minimum set cover over predecessors of $X$ in $G_{\mathrm{occ}}$ via an exact small-set solver (the graphs are small enough that this is tractable); ties are broken by depth-then-lexicographic order.
    \emph{T3 (transparency):} Cast rays from the camera through the volume of $X$; the ground truth is the set of $j$ whose mask intersects any such ray and whose depth is greater than $X$'s back-face depth, estimated from the depth map within $M_X$.
  \item \textbf{Validation gates.} Reject samples where $G_{\mathrm{occ}}$ has fewer than three nodes, where the depth gradient across the occluding chain is below a calibrated minimum (to avoid degenerate cases where ``in front'' is ill-defined), or where multiple objects share a depth band within sensor noise.
\end{enumerate}
The output is a tuple $(\text{image}, \text{prompt}, \text{ground-truth set})$ per task that mirrors the human-produced annotation in structure.

\paragraph{Reflection (T4).}
Reflection annotation is harder to automate than occlusion because the supervision signal is geometric correspondence between a 3D object and a 2D reflection patch on a surface. Our pipeline is:
\begin{enumerate}[leftmargin=1.5em,nosep]
  \item \textbf{Reflective-surface detection.} Run a material-classification head over SAM~3 masks to identify candidate reflective surfaces (polished wood, granite, marble, quartz, glass). Reject scenes with no high-confidence reflective surface.
  \item \textbf{Reflection-region detection.} Within each reflective surface mask, detect candidate reflection patches via a separate detector trained on natural reflections, returning per-patch bounding boxes $\{B_k\}$.
  \item \textbf{3D-to-2D correspondence.} For each detected reflection patch $B_k$, identify the 3D object whose silhouette and pose are consistent with the reflection geometry under a planar-mirror model of the surface. Concretely, mirror each above-surface object's bounding box across the surface plane and rank candidates by geometric overlap with $B_k$. Establish $(B_k \leftrightarrow \text{object}_i)$ pairs above a calibrated overlap threshold; reject ambiguous pairings.
  \item \textbf{Counterfactual variant generation.} Sample a subset $S$ of detected reflection patches to be removed; perform inpainting (a diffusion-based inpainter conditioned on the surrounding surface texture) over $\bigcup_{k \in S} B_k$ to remove those reflections while preserving the rest. The ground-truth answer set is then $\{\text{object}_i : (B_k, i) \notin S\}$, i.e.\ the objects whose reflections survived editing. This mirrors the human protocol described in Appendix~\ref{sec:appendix_imggen}.
  \item \textbf{Validation gates.} Run a perceptual-quality model and a reflection re-detector over the inpainted image; reject samples where (a)~inpainting artifacts are detectable above threshold, or (b)~the re-detector finds reflections at any of the removed locations.
\end{enumerate}
The reflection pipeline is the most failure-prone of the three because it composes three imperfect modules (material classification, reflection detection, planar-mirror correspondence) and an inpainter whose artifacts can themselves leak signal. We expect this task to attract the highest share of human-review flags during the pilot batch.

\paragraph{Volumetric planning (T5, T6).}
The planning tasks are the cleanest to automate because the ground truth is computed by a deterministic search over a 3D-occupancy graph:
\begin{enumerate}[leftmargin=1.5em,nosep]
  \item \textbf{Object volumes.} Use SAM~3 masks plus depth to estimate each object's frustum-projected volume $V_i$ and its $x$-coordinate $x_i$ (lateral position).
  \item \textbf{Target ordering generation.} Sample a target left-to-right ordering by permuting a subset of $\{x_i\}$ such that achieving the target requires either (T5) a unique single pairwise swap or (T6) a multi-step sequence of swaps.
  \item \textbf{Collision-aware swap search.} A pairwise swap $(i, j)$ is \emph{feasible} iff placing $V_i$ at $x_j$ and $V_j$ at $x_i$ produces no overlap with any other $V_k$. For T5, exhaustively enumerate single swaps and retain scenes admitting exactly one feasible solution. For T6, run BFS over swap sequences and retain the shortest valid sequence; tag scenes where no sequence exists as \emph{infeasible}, mirroring the human protocol.
  \item \textbf{Prompt assembly.} Generate the natural-language prompt by populating the T5/T6 template (Appendix~\ref{sec:appendix_prompts}) with the sampled target ordering.
  \item \textbf{Validation gates.} Reject scenes where (a)~object volumes overlap in the original image (a perception failure upstream), or (b)~the search returns multiple equally-shortest solutions for T6 (ambiguous ground truth).
\end{enumerate}

\subsection{Pilot-Batch Validation}
\label{sec:appendix_scaling_validation}

We do not attempt per-sample agreement against the existing 3{,}034-sample release. The scenes admit many valid configurations (occluder counts, surface materials, swap depths, ordering targets) and forcing a pipeline to reproduce one specific human-authored configuration measures stylistic conformity rather than ground-truth quality. The validation objective is instead pipeline reliability over a pilot batch.

\paragraph{Pilot batch.}
The pipeline generates a fresh batch of 3{,}034 samples (matching the size of the human-authored release), with the per-task distribution mirroring the original split. Each sample carries the full output tuple: image, prompt, ground-truth set, and a structured trace of the modules that produced it (detected objects, depth assignments, occlusion graph, validation-gate decisions, etc.). This trace is what makes human review tractable; reviewers see why the pipeline made each call.

\paragraph{Two-layer human review.}
Every pilot sample passes through the same two-stage protocol used for the original release (Appendix~\ref{sec:appendix_annotation}), with one modification: reviewers begin from the pipeline's proposed ground truth rather than from a blank slate, and their job is to verify or flag, not to re-author. Stage~1 (annotator) verifies the image, the prompt, the ground-truth set, and the module trace; flags any inconsistency. Stage~2 (QC reviewer) independently reviews every sample, including all flags raised in Stage~1. Reviewers categorise issues by failure type (object-detection miss, depth-ordering error, reflection-correspondence error, ambiguous swap solution, and so on). Issue counts feed back into pipeline improvement.

\paragraph{Iterative pipeline improvement.}
The pilot is not pass/fail on first review. Flagged issues are aggregated by failure type, the responsible module is improved, and the affected samples (or, where the failure is systemic, the full batch) are regenerated and re-reviewed. The cycle repeats until both review layers sign off on all 3{,}034 samples with no outstanding issues. The number of iterations required is itself a useful signal (a pipeline that needs a single pass is more reliable than one that needs five, even if both arrive at the same final state) and we report the iteration count alongside the released pipeline.

\paragraph{From pilot to scale.}
Once the pilot clears two-layer review with full sign-off, the pipeline is treated as reliable and scaled without further per-sample human review. Released splits at scale carry the pipeline version and a pointer to the pilot review log, so users can audit the basis on which the pipeline was approved. Out of caution, a small rolling spot-check ($\sim$1\% of new samples) is retained as a regression detector for module drift, but it does not gate ingestion.

\subsection{What Scaling Buys, and What it Does Not}
\label{sec:appendix_scaling_scope}

A pilot-validated pipeline buys us the ability to grow the benchmark into regimes the current release cannot support: stratified slices by scene complexity, larger per-task counts that tighten the confidence intervals on near-floor tasks like T4, and rapid generation of held-out test sets for new model releases. It also buys us reproducibility: the pipeline itself is the dataset specification, and any researcher can regenerate or extend the benchmark under the same construction rules.

It does \emph{not} eliminate the need for human judgement, and we are explicit about this. The pilot review is the load-bearing step; everything downstream rests on it. New task families would require their own modules and their own pilot. The pipeline also inherits the biases of its perception modules (a SAM~3 weakness on some object class becomes a benchmark weakness on that class) which the rolling spot-check is designed to surface but cannot eliminate. We view the pipeline less as a replacement for human annotation and more as the path that lets human annotation focus where it has highest marginal value: on the pilot review, on hard cases that survive validation gates, and on the construction of new task families future work will demand.

\end{document}